\documentclass[10pt,letterpaper]{article}

\usepackage{cogsci}
\usepackage{pslatex}
\usepackage{apacite}
\usepackage{graphicx}
\usepackage[export]{adjustbox}

\title{The Causal Frame Problem: An Algorithmic Perspective}
\author{{\large \bf Ardavan S. Nobandegani$^{1,2}$ \quad Ioannis N. Psaromiligkos$^{1}$} \\
\{ardavan.salehinobandegani@mail.mcgill.ca, yannis@ece.mcgill.ca\}\\
$^{1}$Department of Electrical \& Computer Engineering, McGill University\\
$^{2}$Department of Psychology, McGill University}

\newcommand{\bP}{\mathbb P}
\newcommand{\bb}[1]{\textbf{#1}}

\newcommand{\mc}[1]{\mathcal{#1}}
\newcommand{\VV}{\textit Val}

\newcommand{\ind}{\perp\hspace*{-5pt}\perp}
\newcommand{\indd}{\perp\hspace*{-5pt}\perp}

\usepackage{amsmath}
\usepackage{amsfonts}
\usepackage{amssymb}
\usepackage{amsthm}
\usepackage{color}
\usepackage{amsmath}
\usepackage{amsfonts}
\usepackage{amssymb}
\usepackage{bm}

\begin{document}
\maketitle
\begin{abstract}
The Frame Problem (FP) is a puzzle in philosophy of mind and epistemology, articulated by the Stanford Encyclopedia of Philosophy as follows: \emph{``How do we account for our apparent ability to make decisions on the basis only of what is relevant to an ongoing situation without having explicitly to consider all that is not relevant?"} In this work, we focus on the \emph{causal} variant of the FP, the Causal Frame Problem (CFP). Assuming that a reasoner's mental causal model can be (implicitly) represented by a causal Bayes net, we first introduce a notion called Potential Level (PL). PL, in essence, encodes the relative position of a node with respect to its neighbors in a causal Bayes net. Drawing on the psychological literature on causal judgment, we substantiate the claim that PL may bear on how \emph{time} is encoded in the mind. Using PL, we propose an inference framework, called the PL-based Inference Framework (PLIF), which permits a boundedly-rational approach to the CFP to be formally articulated at Marr's algorithmic level of analysis. We show that our proposed framework, PLIF, is consistent with a wide range of findings in causal judgment literature, and that PL and PLIF make a number of predictions, some of which are already supported by existing findings. 

\end{abstract}

\section{Introduction}
At the core of any decision-making or reasoning task, resides an innocent-looking yet challenging question: Given an inconceivably large body of knowledge available to the reasoner, what constitutes the relevant for the task and what the irrelevant? The question, as it is posed, echoes the well-known Frame Problem (FP) in epistemology and philosophy of mind, articulated by Glymour (1987) as follows: \emph{``Given an enormous amount of stuff, and some task to be done using some of the stuff, what is the relevant stuff for the task?"} Fodor (1987) comments: \emph{``The frame problem goes very deep; it goes as deep as the analysis of rationality."}

The question posed above perfectly captures what is really at the core of the FP, yet, it may suggest an unsatisfying approach to the FP at the algorithmic level of analysis (Marr, 1982). Indeed, the question may suggest the following two-step methodology: In the first step, out of all the body of knowledge available to the reasoner (termed, the model), she has to identify what is relevant to the task (termed, the relevant submodel); it is \emph{only then} that she advances to the second step by performing \emph{reasoning} or \emph{inference} on the identified submodel. There is something fundamentally wrong with this methodology (which we term, \emph{sequential} approach to reasoning) which bears on the following understanding: The relevant submodel, i.e., the portion of the reasoner's knowledge deemed relevant to the task, oftentimes is so enormous (or even infinitely large) that the reasoner---inevitably bounded in time and computational resources---would never get to the second step, had she adhered to such a methodology. In other words, in line with the notion of bounded rationality (Simon, 1957), a boundedly-rational reasoner must have the option, if need be, to merely consult a fraction of the potentially large---if not infinitely so---relevant submodel.

Recent work by \citeA{icard2015} elegantly promotes this insight when they write: \emph{``Somehow the mind must focus in on some ``submodel" of the ``full" model (including all possibly relevant variables) that suffices for the task at hand and is not too costly to use."}\footnote{In an informative example on Hidden Markov Models (HMMs), Icard \& Goodman (2015) present a setting wherein the relevant submodel is infinitely large---an example which makes it pronounced what is wrong with the sequential approach stated earlier.} They then ask the following question: \emph{``what kind of simpler model should a reasoner consult for a given task?"} This is an inspiring question hinting to an interesting line of inquiry as to how to formally articulate a boundedly-rational approach to the FP at Marr's algorithmic level of analysis (1982).

In this work, we focus on the causal variant of the FP, the Causal Frame Problem (CFP), stated as follows: Upon being presented with a causal query, how does the reasoner manage to attend to her causal knowledge relevant to the derivation of the query while rightfully dismissing the irrelevant? We adopt Causal Bayesian Networks (CBNs) (Pearl, 1988; Gopnik et al., 2004, \emph{inter alia}) as a normative model to represent how the reasoner's \emph{internal} causal model of the world is structured (i.e., reasoner's mental model). First, we introduce the notion of Potential Level (PL). PL, in essence, encodes the relative position of a node (representing a propositional variable or a concept) with respect to its neighbors in a CBN. Drawing on the psychological literature on causal judgment, we substantiate the claim that PL may bear on how \emph{time} is encoded in the mind. Equipped with PL, we embark on investigating the CFP at Marr's algorithmic level of analysis. We propose an inference framework, termed PL-based Inference Framework (PLIF), which aims at empowering the boundedly-rational reasoner to consult (or retrieve\footnote{The terms ``consult" and ``retrieve" will be used interchangeably. We elaborate on the rationale behind that in Sec. \ref{sec_discussion}, where we connect our work to Long Term Memory and Working Memory.}) parts of the underlying CBN deemed relevant for the derivation of the posed query (the relevant submodel) in a \emph{local}, \emph{bottom-up} fashion until the submodel is fully retrieved. PLIF allows the reasoner to carry out inference at \emph{intermediate} stages of the retrieval process over the thus-far retrieved parts, thereby obtaining lower and upper bounds on the posed causal query. We show, in the Discussion section, that our proposed framework, PLIF, is consistent with a wide range of findings in causal judgment literature, and that PL and PLIF make a number of predictions, some of which are already supported by the findings in the psychology literature.

In their work, Icard and Goodman (2015) articulate a boundedly-rational approach to the CFP at Marr's computational level of analysis, which, as they point out, is from a ``god's eye" point of view. In sharp contrast, our proposed framework PLIF is \emph{not} from a ``god's eye" point of view and hence could be regarded, potentially, as a psychologically plausible proposal at Marr's algorithmic level of analysis as to how the mind both retrieves and, at the same time, carries out inference over the retrieved submodel to derive bounds on a causal query. We term this \emph{concurrent} approach to reasoning, as opposed to the flawed sequential approach stated earlier.\footnote{We elaborate more on this in the Discussion section.} The retrieval process progresses in a local, bottom-up fashion, hence the submodel is retrieved \emph{incrementally}, in a \emph{nested} manner.\footnote{The term ``nested" implies that the thus-far retrieved submodel is subsumed by every later submodel (should the reasoner proceeds with the retrieval process).} Our analysis (Sec. \ref{sec_case_study}) confirms Icard and Goodman's insight (2015) that even in the extreme case of having an infinitely large relevant submodel, the portion of which the reasoner has to consult so as to obtain a ``sufficiently good" answer to a query could indeed be very small.

\section{Potential Level and Time}
\label{sec:PL_vs_time}
Before proceeding further, let us introduce some preliminary notations. Random Variables (RVs) are denoted by lower-case bold-faced letters, e.g., $\bb{x}$, and their realizations by non-bold lower-case letters, e.g., $x$. Likewise, sets of RVs are denoted by upper-case bold-faced letters, e.g., $\bb X$, and their corresponding realizations by upper-case non-bold letters, e.g., $X$. $\textit{Val}(\cdot)$ denotes the set of possible values a random quantity can take on. Random quantities are assumed to be discrete unless stated otherwise.
The joint probability distribution over $\bb{x}_1,\cdots,\bb{x}_n$ is denoted by $\bP(\bb{x}_1,\cdots,\bb{x}_n)$. We will use the notation $\bb{x}_{1:n}$ to denote the sequence of $n$ RVs $\bb{x}_1,\cdots,\bb{x}_n$, hence $\bP(\bb{x}_1,\cdots,\bb{x}_n)=\bP(\bb{x}_{1:n})$. The terms ``node" and ``variable" will be used interchangeably throughout. To simplify presentation, we adopt the following notation: We denote the  probability $\bP(\bb{x}=x)$ by $\bP(x)$ for some RV $\bb{x}$ and its realization $x\in \textit{Val}(\bb{x})$. For conditional probabilities, we will use the notation $\bP(x|y)$ instead of $\bP(\bb{x}={x}|\bb{y}={y})$. Likewise, $\bP(X|Y)=\bP(\bb X= X|\bb Y= Y)$ for $X \in \VV(\bb X)$ and $Y \in \VV(\bb Y)$. A generic conditional independence relationship is denoted by $(\bb A \indd \bb B|\bb C)$ where $\bb A, \bb B$, and $\bb C$ represent three mutually disjoint sets of variables belonging to a CBN. Furthermore, throughout the paper, we assume that $\epsilon$ is some negligibly small positive real-valued quantity. Whenever we subtract $\epsilon$ from a quantity, we simply imply a quantity less than but arbitrarily close to the original quantity. The rationale behind adopting such a notation will become clearer in Sec. \ref{sec_PLIF_main}.

Before formally introducing the notion of PL (unavoidably, with some mathematical jargon), we articulate in simple terms what the idea behind PL is. PL simply induces a \emph{chronological order} on the nodes of a CBN, allowing the reasoner to encode the timing between cause and effect.\footnote{More precisely, PL induces a topological order on the nodes of a CBN, with temporal interpretations suggested in Def. 1.} As we will see, PL plays an important role in guiding the retrieval process used in our proposed framework. Next, PL is formally defined, followed by two clarifying examples.

\textbf{Def. 1. (Potential Level (PL))} Let $par(\bb x)$ and $child(\bb x)$ denote, respectively, the sets of parents (i.e., immediate causes) and children (i.e., immediate effects) of $\bb x$. Also let $T_0\in\mathbb{R}\cup\{-\infty\}$. The PL of $\bb x$, denoted by $p_l(\bb x)$, is defined as follows: (i) If $par(\bb x)=\varnothing$, $p_l(\bb x)=T_0$, and (ii) If $par(\bb x)\neq\varnothing$, $p_l(\bb x)$ is a real-valued quantity selected from the interval $(\max_{\bb y\in par(\bb x)}p_l(\bb y),\min_{\bb z\in child(\bb x)}p_l(\bb z))$ such that $p_l(\bb x)-\max_{\bb y\in par(\bb x)}p_l(\bb y)$ indicates the amount of time which elapses between intervening simultaneously on all the RVs in $par(\bb x)$ (i.e., $do(par(\bb x)=par_x)$) and $\bb x$ taking its value $x$ in accord with the distribution $\bP(x|par_x)$. If $child(\bb x)=\varnothing$, substitute the upper bound of the given interval by $+\infty$. \hfill $\blacksquare$

Parameter $T_0$ symbolizes the origin of time, as perceived by the reasoner. $T_0=0$ is a natural choice, unless the reasoner believes that time continues indefinitely into the past, in which case $T_0=-\infty$. The next two examples further clarify the idea behind PL. In both examples we assume $T_0=0$.

\begin{figure}[h!]
\centering
\includegraphics[width=0.4\textwidth]{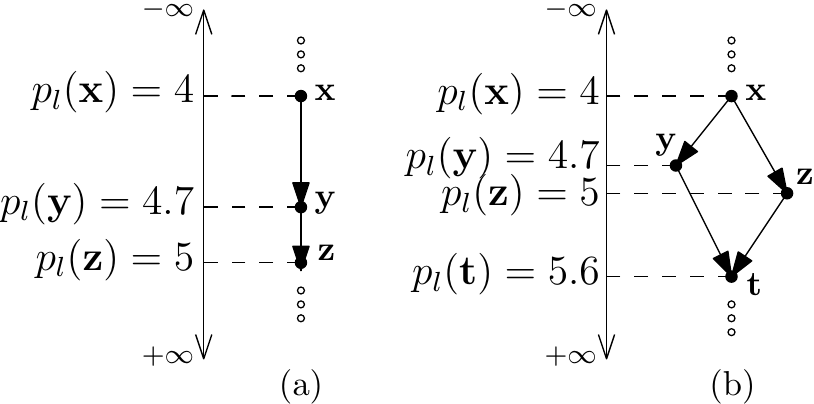}
\caption{Relation between PL and time: Example.}
\label{fig_example_1}
\end{figure}

For the first example, let us consider the CBN depicted in Fig. \ref{fig_example_1}(a) containing the RVs $\bb x, \bb y,$ and $\bb z$ with $p_l(\bb x)=4, p_l(\bb y)=4.7,$ and $p_l(\bb z)=5$. According to Def. 1, the given PLs can be construed in terms of the relative time between the occurrence of cause and effect as articulated next. Upon intervening on $\bb x$ (i.e., $do(\bb x=x)$), after the elapse of $p_l(\bb y)- p_l(\bb x)=0.7$ units of time, the RV $\bb y$ takes its value $y$ in accord with the distribution $\bP(y|x)$. Likewise, upon intervening on $\bb y$ (i.e., $do(\bb y=y)$), after the elapse of $p_l(\bb z)- p_l(\bb y)=0.3$ units of time, $\bb z$ takes its value $z$ according to $\bP(z|y)$.

For the second example, consider the CBN depicted in Fig. \ref{fig_example_1}(b) containing the RVs $\bb x, \bb y, \bb z$, and $\bb t$ with $p_l(\bb x)=4, p_l(\bb y)=4.7, p_l(\bb z)=5$, and $p_l(\bb t)=5.6$. Upon intervening on $\bb x$ (i.e., $do(\bb x=x)$) the following happens: (i) after the elapse of $p_l(\bb y)- p_l(\bb x)=0.7$ units of time, $\bb y$ takes its value $y$ according to $\bP(y|x)$, and (ii) after the elapse of $p_l(\bb z)- p_l(\bb x)=1$ unit of time, $\bb z$ takes its value $z$ according to $\bP(z|x)$. Also, upon intervening simultaneously on RVs $\bb y, \bb z$ (i.e., $do(\bb y=y,\bb z=z)$), after the elapse of $p_l(\bb t)-\max_{\bb r\in par(\bb t)}p_l(\bb r)=0.6$ units of time, $\bb t$ takes its value $t$ according to $\bP(t|y,z)$.

In sum, the notion of PL bears on the underlying time-grid upon which a CBN is constructed, and adheres to Hume's principle of temporal precedence of cause to effect \cite{hume1975inquiry}. A growing body of work in psychology literature corroborates Hume's centuries-old insight, suggesting that the timing and temporal order between events strongly influences how humans induce causal structure over them \cite{bramley2014order,lagnado2006time}. The introduced notion of PL is based on the following hypothesis: \emph{When learning the underlying causal structure of a domain, humans may as well encode the temporal patterns (or some estimates thereof) on which they rely to infer the causal structure.} This hypothesis is supported by recent findings suggesting that people have expectations about the delay length between cause and effect \cite{greville2010temporal,buehner2004abolishing,schlottmann1999seeing}. It is worth noting that we could have defined PL in terms of relative \emph{expected time} between cause and effect, rather than relative absolute time. Under such an interpretation, the time which elapses between the intervention on a cause and the occurrence of its effect would be modeled by a probability distribution, and PL would be defined in terms of the expected value of that distribution. Our proposed framework, PLIF, is indifferent as to whether PL should be construed in terms of absolute or expected time. \citeA{greville2010temporal} show that causal relations with fixed temporal intervals are consistently judged as stronger compared to those with variable temporal intervals. This finding, therefore, seems to suggest that people expect, to a greater extent, fixed temporal intervals between cause and effect, rather than variable ones---an interpretation which, at least to a first approximation, favors construing PL in terms of relative absolute time (see Def. 1).\footnote{There are cases, however, that, despite the precedence of cause to effect, quantifying the amount of time between their occurrences may bear no meaning, e.g., when dealing with hypothetical constructs. In such cases, PL should be simply construed as a topological ordering. From a purely computational perspective, PL is a generalization of \emph{topological sorting} in computer science.}

\section{Informative Example}
\label{Sec_toy_example_I}
To develop our intuition, and before formally articulating our proposed framework, let us present a simple yet informative example which demonstrates: (i) how the retrieval process can be carried out in a local, bottom-up fashion, allowing for retrieving the relevant submodel incrementally, and (ii) how adopting PL allows the reasoner to obtain bounds on a given causal query at intermediate stages of the retrieval process.

\begin{figure}[h!]
\centering
\includegraphics[width=0.4\textwidth]{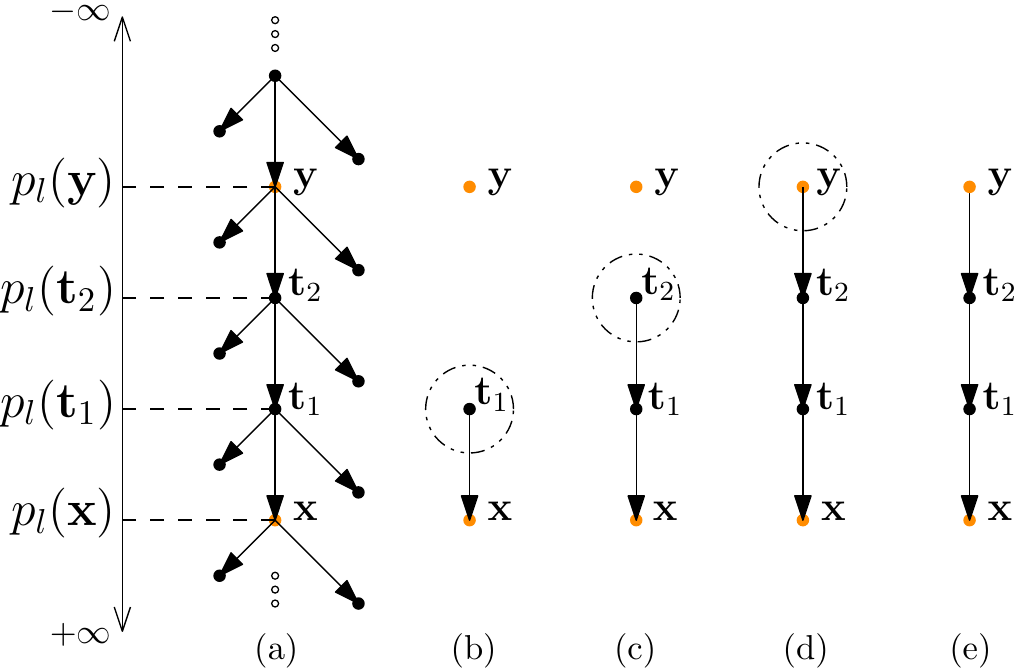}
\caption{Example. Query variables are shown in orange.}
\label{fig_motive}
\end{figure}

Let us assume that the posed causal query is $\bP(x|y)$ where $\bb x, \bb y$ are two RVs in the CBN depicted in Fig. \ref{fig_motive}(a) with PLs $p_l(\bb x),p_l(\bb y)$, and let $p_l(\bb x)>p_l(\bb y)$. The relevant information for the derivation of the posed query (i.e., the relevant submodel) is depicted in Fig. \ref{fig_motive}(e).

Starting from the target RV $\bb x$ in the original CBN (Fig. \ref{fig_motive}(a)) and moving one step backwards,\footnote{Taking one step backwards from variable $\bm q$ amounts to retrieving all the parents of $\bm q$.} $\bb t_1$ is reached (Fig. \ref{fig_motive}(b)). Since $p_l(\bb y)< p_l(\bb t_1)$, $\bb y$ must be a non-descendant of $\bb t_1$, and therefore, of $\bb x$. Hence, conditioning on $\bb t_1$ $d$-separates $\bb x$ from $\bb y$ \cite{pearl2014probabilistic}, yielding $(\bb x\indd \bb y|\bb t_1)$. Thus $\bP(x|y)=\sum_{t_1\in\bb Val(t_1)}\bP(x|y,t_1)\bP(t_1|y)=\sum_{t_1\in\bb Val(t_1)}\bP(x|t_1)\bP(t_1|y)$ implying: $\min_{t_1\in Val(\bb t_1)}\bP(x|t_1)\leq \bP(x|y)\leq \max_{t_1\in Val(\bb t_1)}\bP(x|t_1)$. It is crucial to note that the given bounds can be computed using the information thus-far retrieved, i.e., the information encoded in the submodel shown in Fig. \ref{fig_motive}(b). Taking a step backwards from $\bb t_1$, $\bb t_2$ is reached (Fig. \ref{fig_motive}(c)). Using a similar line of reasoning to the one presented for $\bb t_1$, having $p_l(\bb y)< p_l(\bb t_2)$  ensures $(\bb x\indd \bb y|\bb t_2)$. Therefore, the following bounds on the posed query can be derived, which, crucially, can be computed using the information thus-far retrieved:  $\min_{t_2\in Val(\bb t_2)}\bP(x|t_2)\leq \bP(x|y) \leq \max_{t_2\in Val(\bb t_2)}\bP(x|t_2)$. It is straightforward to show that the bounds derived in terms of $\bb t_2$ are tighter than the bounds derived in terms of $\bb t_1$.\footnote{Here we are implicitly making the assumption that the CPDs involved in the parameterization of the underlying CBN are non-degenerate. Dropping this assumption yields the following result: The bounds derived in terms of $\bb t_2$ are equally-tight or tighter than the bounds derived in terms of $\bb t_1$.} Finally, taking one step backward from $\bb t_2$, $\bb y$ is reached (Fig. \ref{fig_motive}(d)) and the exact value for $\bP(x|y)$ can be derived, again using  only the submodel thus-far retrieved (Fig. \ref{fig_motive}(d)).

We are now well-positioned to present our proposed framework, PLIF.

\section{PL-based Inference Framework (PLIF)}
\label{sec_PLIF_main}
In this section, we intend to elaborate on how, equipped with the notion of PL, a generic causal query of the form\footnote{We do not consider interventions in this work. However, with some modifications, the presented analysis/results can be extended to handle a generic causal query of the form $\bP(\bb O=O| \bb E=E, do(\bb Z=Z))$ where $\bb Z$ denotes the set of intervened variables.} $\bP(\bb O=O|\bb E=E)$ can be derived where  $\bb O$ and $\bb E$ denote, respectively, the disjoint sets of target (or \emph{objective}) and observed (or \emph{evidence}) variables. In other words, we intend to formalize how inference over a CBN whose nodes are endowed with PL as an attribute should be carried out. Before we present the main result, a few definitions are in order.

\textbf{Def. 2. (Critical Potential Level (CPL))} The target variable with the least PL is denoted by $\bb o^\ast$ and its PL is referred to as the CPL. More formally, $p_l^{\ast}:\triangleq\min_{\bb o\in \bb O} p_l(\bb o)$ and $\bb o^\ast:\triangleq\arg\min_{\bb o\in \bb O} p_l(\bb o)$. E.g., for the setting given in Fig. \ref{fig_motive}(a), $\bb o^\ast=\bb x$, and $p_l^{\ast}=p_l(\bb x)$. Viewed through the lens of time, $\bb o^\ast$ is the furthest target variable into the past, with PL $p_l^{\ast}$.

There are two possibilities: (a) $p_l^\ast>T_0$, or (b) $p_l^\ast=T_0$, with $T_0$ denoting the origin of time; cf. Sec. \ref{sec:PL_vs_time}. In the sequel we assume that (a) holds. For a discussion on the special case (b), the reader is referred to the Supplementary Information.

\textbf{Def. 3. (Inference Threshold (IT) and IT Root Set (IT-RS))}  To any real-valued quantity, $\mc T$, corresponds a unique set, $\bb R_{\mc T}$, obtained as follows: Start at every variable $\bb x\in \bb O\cup\bb E$ with PL $\geq \mc T$ and backtrack along all paths terminating at $\bb x$. Backtracking along each path stops as soon as a node with PL less than $\mc T$ is encountered. Such nodes, together, compose the set $\bb R_{\mc T}$. It follows from the definition that: $\max_{\bb t\in\bb R_{\mc T}}p_l(\bb t)<\mc T$. $\mc T$ and $\bb R_{\mc T}$ are termed, respectively, Inference Threshold (IT) and the IT Root Set (IT-RS) for $\mc T$.

For example, the set of variables circled at the stages depicted in Figs. \ref{fig_motive}(b-d) are, the IT-RSs for $\mc T=p_l(\bb x)-\epsilon$, $\mc T=p_l(\bb t_1)-\epsilon$, and $\mc T=p_l(\bb t_2)-\epsilon$, respectively. Note that instead of, say $\mc T=p_l(\bb x)-\epsilon$, we could have said: for any $\mc T\in(p_l(\bb t_1),p_l(\bb x))$. However, expressing ITs in terms of $\epsilon$ liberates us from having to express them in terms of intervals thereby simplifying the exposition in the sole hope that the reader finds it easier to follow the work. We would like to emphasize that the adopted notation should \emph{not} be construed as implying that the assignment of values to ITs is such a sensitive task that everything would have collapsed, had IT not been chosen in such a fine-tuned manner. To recap, in simple terms, $\mc T$ bears on how far into the past a reasoner is consulting her mental model in the process of answering a query, and $\bb R_{\mc T}$ characterizes the furthest-into-the-past concepts entertained by the reasoner in that process.

Next, we formally present the main idea behind PLIF, followed by its interpretation in simple terms.

\textbf{Lemma 1.} For any chosen IT $\mc T<p_l^\ast$ and its corresponding $\bb R_{\mc T}$, define $\bb S:\triangleq\bb R_{\mc T}\setminus\bb E$. Then the following holds:
\begin{eqnarray}
\label{eq_PLIF}
\min_{S\in Val(\bb S)}\bP(O|S,E)\leq \bP(O|E) \leq\max_{S\in Val(\bb S)}\bP(O|S,E).
\end{eqnarray}
Crucially, the provided bounds can be computed using the information encoded in the submodel retrieved in the very process of obtaining the $\bb R_{\mc T}$. \hfill $\square$

For a formal proof of Lemma 1, the reader is referred to the Supplementary Information. Mathematical jargon aside, the message of Lemma 1 is quite simple: For any chosen inference threshold $\mc T$ which is further into the past than $\bb o^\ast$, Lemma 1 ensures that the reasoner can condition on $\bb S$ and obtain the reported lower and upper bounds on the query by using \emph{only} the information encoded in the retrieved submodel.

It is natural to ask under what conditions the exact value to the posed query can be derived using the thus-far retrieved submodel (i.e., the submodel obtained during the identification of $\bb R_{\mc T}$). The following remark bears on that.

\textbf{Remark 1.} If for IT $\mc T$, $\bb R_{\mc T}$ satisfies either: (i) $\bb R_{\mc T}\subseteq \bb E$, or (ii) for all $\bb r\in\bb R_{\mc T},\hspace*{3pt} p_l(\bb r)=T_0$, and $\min_{\bb e\in \bb E}p_l(\bb e)> \mc T$, or (iii) the lower and upper bound given in (\ref{eq_PLIF}) are identical, then the exact value of the posed query can be derived using the submodel retrieved in the process of obtaining $\bb R_{\mc T}$. Fig. \ref{fig_motive}(d) shows a setting wherein conditions (i) and (iii) are both met.

The rationale behind Remark 1 is provided in the Supplementary Information.

\subsection{Case Study}
\label{sec_case_study}
Next, we intend to cast the Hidden Markov Model (HMM) studied in (Icard \& Goodman, 2015, p. 2) into our  framework. 
\begin{figure}[h!]
\centering
\includegraphics[trim = 0mm -4mm 5mm 0mm, clip, width=3.407cm]{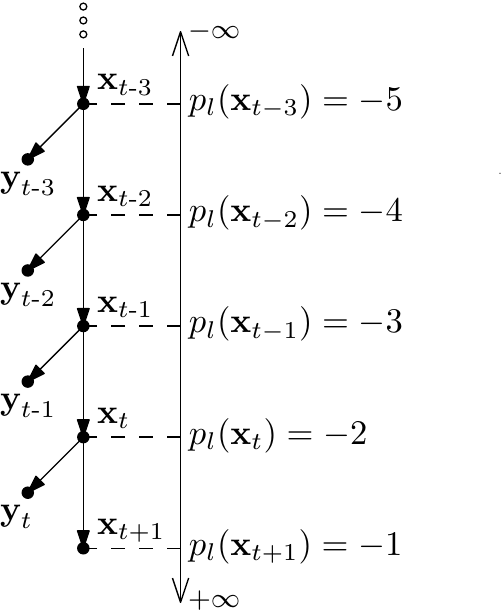}
\includegraphics[trim = 52mm 86mm 55mm 92mm, clip, width=0.286\textwidth]{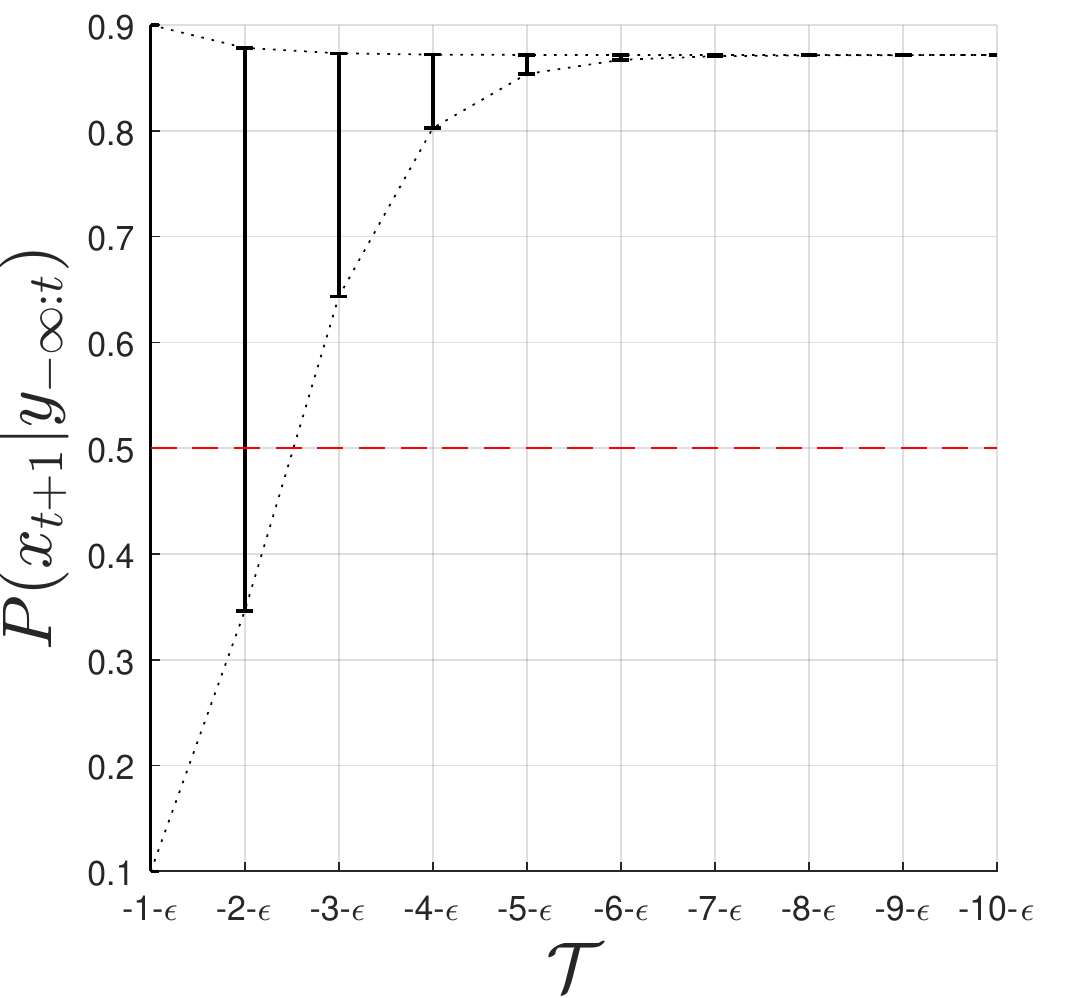}
\caption{Left: The infinite-sized HMM discussed in (Icard \& Goodman, 2015) with parameterization: $\bP(x_{t+1}|x_t)=\bP(\bar{x}_{t+1}|\bar{x}_t)=0.9$, and $\bP(y_t|x_t)=\bP(\bar{y}_t|\bar{x}_t)=0.8$. Right: Applying PLIF on the HMM shown in left. Vertical and horizontal axes denote, respectively, the value of the posed query $\bP(x_{t+1}|y_{-\infty:t})$ and the adopted IT $\mc T$. The vertical bars depict the intervals within which the query lies due to Lemma 1. The dotted curves---which connect the lower and upper bounds of the intervals---show how the intervals shrink as IT $\mc T$ decreases.}
\label{fig_HMM_Icard_Goodman}
\end{figure}
The setting is shown in Fig. \ref{fig_HMM_Icard_Goodman}(left). We adhere to the same parametrization and query adopted therein. All RVs in this section are binary, taking on values from the set $\{0,1\}$; $\bb x=x$ indicates the event wherein $\bb x$ takes the value 1, and $\bb x=\bar{x}$ implies the event wherein $\bb x$ takes the value 0. We assume $p_l(\bb x_{t+i})=i-2$.\footnote{Note that the trend of the upper- and lower-bound curve as well as the size of the intervals shown in Fig. \ref{fig_HMM_Icard_Goodman}(right) are insensitive with regard to the choice of PLs for variables $\{\bb x_{t-i}\}_{i=-1}^{+\infty}$.} We should note that the assignment of the PLs for the variables in $\{\bb y_{t-i}\}_{i=0}^{+\infty}$ does not affect the presented results in any way. The query of interest is $\bP(x_{t+1}|y_{-\infty:t})$. Notice that after performing three steps of the sort discussed in the example presented in Sec. \ref{Sec_toy_example_I} (for the IT $\mc T=-3-\epsilon$), the lower bound on the posed query exceeds 0.5 (shown by the red dashed line in Fig. \ref{fig_HMM_Icard_Goodman}(right)). This observation has the following intriguing implication. Assume, for the sake of argument, that we were presented with the following Maximum A-Posterior (MAP) inference problem: Upon observing all the variables in $\{\bb y_{t-i}\}_{i=0}^{+\infty}$ taking on the value 1, what would be the most likely state for the variable $\bb x_{t+1}$? Interestingly, we would be able to answer this MAP inference problem simply after three backward moves (corresponding to the IT $\mc T=-3-\epsilon$). In Fig. \ref{fig_HMM_Icard_Goodman}(right), the intervals within which the posed query falls (due to Lemma 1) in terms of the adopted IT $\mc T$ are depicted.

Our analysis confirms Icard and Goodman's insight (2015) that even in the extreme case of having infinite-sized relevant submodel (Fig. \ref{fig_HMM_Icard_Goodman}(left)), the portion of which the reasoner has to consult so as to obtain a ``sufficiently good" answer to the posed query could happen to be very small (Fig. \ref{fig_HMM_Icard_Goodman}(right)).

\section{Discussion}
\label{sec_discussion}
To our knowledge, PLIF is the first inference framework proposed that capitalizes on \emph{time} to constrain the scope of causal reasoning over CBNs, where the term scope refers to the portion of a CBN on which inference is carried out. PLIF does not restrict itself to any particular inference scheme. The claim of PLIF is that inference should be confined within and carried out over retrieved submodels of the kind suggested by Lemma 1 so as to obtain the reported bounds therein. In this light, PLIF can accommodate all sorts of inference schemes, including Belief Propagation (BP), and sample-based inference methods using Markov Chain Monte Carlo (MCMC), as two prominent classes of inference schemes proposed in the literature.\footnote{{MCMC-based methods have been successful in simulating important aspects of a wide range of cognitive phenomena, and giving accounts for many cognitive biases; cf. \cite{sanborn2016bayesian}. Also, work in theoretical neuroscience has suggested  mechanisms for how BP and MCMC-based methods could be realized in neural circuits; cf. \cite{gershman2016complex,lochmann2011neural}.}} For example, to cast BP into PLIF amounts to restricting BP's message-passing within submodels of the kind suggested by Lemma 1. In other words, assuming that BP is to be adopted as the inference scheme, upon being presented with a causal query, an IT according to Lemma 1 will be selected---at the \emph{meta-level}---by the reasoner and the corresponding submodel, as suggested by Lemma 1, will be retrieved, over which inference will be carried out using BP. This will lead to obtaining lower and upper bounds on the query, as reported in Lemma 1. If time permits, the reasoner builds up \emph{incrementally} on the thus-far retrieved submodel so as to obtain tighter bounds on the query.\footnote{The very property that the submodel gets constructed incrementally in a nested fashion guarantees that the obtained lower and upper bounds get tighter as the reasoner adopts smaller ITs; cf. Fig. \ref{fig_HMM_Icard_Goodman}(left).} MCMC-based inference methods can be cast, in a similar fashion, into PLIF.

The problem of what parts of a CBN are relevant and what are irrelevant for a given query, according to (Geiger, Verma, \& Pearl, 1989), was first addressed by Shachter (1988). The approaches proposed for identifying the relevant submodel for a given query fall into two broad categories (cf. (Mahoney \& Laskey, 1998) and references therein): (i) top-down approaches, and (ii) bottom-up approaches. Top-down approaches start with the full knowledge of the underlying CBN and, depending on the posed query, gradually \emph{prune} the irrelevant parts of the CBN. In this respect, top-down approaches are inevitably from ``god's eye" point of view---a characteristic which undermines their cognitive-plausibility. Bottom-up approaches, on the other hand, start at the variables involved in the posed query and move backwards till the \emph{boundaries} of the underlying CBN are finally reached, only then they start to prune the parts of the constructed submodel---if any---which can be safely removed without jeopardizing the exact computation of the posed query. It is important to note that bottom-up approaches cannot stop at \emph{intermediate} steps during the backward move and run inference on the thus-far constructed submodel without running the risk of compromising some of the (in)dependence relations structurally encoded in the CBN, which would yield erroneous inferences. This observation is due to the fact that there exists no local signal revealing how the thus-far retrieved nodes are positioned relative to each other and to the to-be-retrieved nodes---a shortcoming circumvented in the case of PLIF by introducing PL. Another pitfall shared by both top-down and bottom-up approaches is their \emph{sequential} methodology towards the task of inference, according to which the relevant submodel for the posed query should be first constructed, and only then inference is carried out to compute the posed query.\footnote{The computation can be carried out to obtain either the exact value or simply an approximation to the query. Nonetheless, what both top-down and bottom-up approaches agree on is that the relevant submodel is to be first identified, should the reasoner intend to compute exactly or approximately the posed query.} On the contrary, PLIF submits to what we call the \emph{concurrent} approach to reasoning, whereby retrieval and inference take place \emph{in tandem}. The HMM example analyzed in Sec. \ref{sec_case_study}, shows the efficacy of the concurrent approach.

Work on causal judgment provides support for the so-called alternative neglect, according to which subjects tend to neglect alternative causes to a much greater extent in predictive reasoning than in diagnostic reasoning \cite{fernbach2013cognitive,fernbach2011asymmetries}. Alternative neglect, therefore, implies that subjects would tend to ignore parts of the relevant submodel while constructing it. Recent findings, however, seem to cast doubt on alternative neglect \cite{cummins2014impact,meder2014structure}. Meder et al. (2014), Experiment 1 demonstrates that subjects appropriately take into account alternative causes in predictive reasoning. Also, Cummins (2014) substantiates a two-part explanation of alternative neglect according to which: (i) subjects interpret predictive queries as requests to estimate the probability of the effect when only the focal cause is present, an interpretation which renders alternative causes irrelevant, and (ii) the influence of inhabitory causes (i.e., disablers) on predictive judgment is underestimated, and this underestimation is incorrectly interpreted as neglecting of alternative causes. Cummins (2014), Experiment 2 shows that when predictive inference is queried in a manner that  more accurately expresses the meaning of noisy-OR Bayes net (i.e., the normative model adopted by \citeA{fernbach2011asymmetries}) likelihood estimates approached normative estimates. \citeA{cummins2014impact}, Experiment 4 shows that the impact of disablers on predictive judgments is far greater than that of alternative causes, while having little impact on diagnostic judgments. PLIF commits to the retrieval of enablers as well as disablers. As mentioned earlier, PLIF abstracts away from the inference algorithm operating on the retrieved submodel, and, hence, leaves it to the inference algorithm to decide how the retrieved enablers and disablers  should be integrated. In this light, PLIF is consistent with the results of Experiment 4.

In an attempt to explain violations of screening-off reported in the literature, \citeA{park2013mechanistic} find strong support for the contradiction hypothesis followed by the mediating mechanism hypothesis, and finally conclude that people do conform to screening-off once the causal structure they are using is correctly specified. PLIF is consistent with these findings, as it adheres to the assumption that reasoners carry out inference on their \emph{internal} causal model (including all possible mediating variables and disablers), not the potentially incomplete one presented in the cover story; see also \cite{Rehder2015,sloman2015causality}.

Experiment 5 in \cite{cummins2014impact}, consistent with \cite{fernbach2013cognitive}, shows that causal judgments are strongly influenced by memory retrieval/activation processes, and that both number of disablers and order of disabler retrieval matter in causal judgments. These findings suggest that the CFP and memory retrieval/activation are intimately linked. In that light, next, we intend to elaborate on the rationale behind adopting the term ``retrieve" and using it interchangeably with the term ``consult" throughout the paper; this is where we relate PLIF to the concepts of Long Term Memory (LTM) and Working Memory (WM) in psychology and neurophysiology.  Next, we elaborate on how PLIF could be interpreted through the lenses of two influential models of WM, namely, Baddeley and Hitch's (1974) Multi-component model of WM (M-WM) and Ericsson and Kintsch's Long-term Working Memory (LTWM) model (1995). The M-WM postulates that \emph{``long-term information is downloaded into a separate temporary store, rather than simply activated in LTM"}, a mechanism which permits WM to \emph{``manipulate and create new representations, rather than simply activating old memories"} (Baddeley, 2003). Interpreting PLIF through the lens of the M-WM model amounts to the value for IT being chosen (and, if time permits, updated so as to obtain tighter bounds) by the central executive in the M-WM and the submodel being incrementally ``retrieved" from LTM into M-WM's episodic buffer. Interpreting PLIF through the lens of the LTWM model amounts to having no retrieval from LTM into WM and the submodel suggested by Lemma 1 being merely ``activated in LTM" and, in that sense, being simply ``consulted" in LTM. In sum, PLIF is compatible with both of the narratives provided by the M-WM and LTWM models.

A number of predictions follow from PL and PLIF. For instance, PLIF makes the following prediction: Prompted with a predictive or a diagnostic query (i.e., $\bP(\bb e|\bb c)$ and $\bP(\bb c|\bb e)$, respectively), subjects should not retrieve any of the effects of $\bb e$. Introspectively, this prediction seems plausible, and can be tested, using a similar approach to \cite{cummins2014impact,de2003inference}, by asking subjects to ``think aloud" while engaging in predictive or diagnostic reasoning. Also, PL yields the following prediction: Upon intervening on cause $\bb c$, subjects should be sensitive to \emph{when} effect $\bb e$ will occur, even in settings where they are not particularly instructed to attend to such temporal patterns. This prediction is supported by recent findings suggesting that people do have expectations about the delay length between cause and effect \cite{greville2010temporal,buehner2004abolishing}.

There is a growing acknowledgment in the literature that, not only time and causality are intimately linked, but that they \emph{mutually constrain} each other in human cognition \cite{buehner2014time}. In line with this view, we see our work also as an attempt to formally articulate how time could guide and constrain causal reasoning in  cognition. While many questions remain open, we hope to have made some progress towards better understanding of the CFP at the algorithmic level.

\section*{Acknowledgments}
We are grateful to Thomas Icard for valuable discussions. We would also like to thank Marcel Montrey and Peter Helfer for helpful comments on an earlier draft of this work. This work was supported in part by the Natural Sciences and Engineering Research Council of Canada under grant RGPIN 262017.

\nocite{mahoney1998constructing}
\nocite{fodor1987modules}
\nocite{icard2015}
\nocite{gopnik2004theory}
\nocite{shachter1988probabilistic}
\nocite{baddeley2003working}
\nocite{ericsson1995long}
\nocite{pearl2014probabilistic}
\nocite{geiger1989d}
\nocite{simon1957models}
\nocite{marr1982vision}
\nocite{glymour1987android}
\nocite{baddeley1974working}
\nocite{Rehder2015}

\renewcommand\bibliographytypesize{\footnotesize}
\bibliographystyle{apacite}
\setlength{\bibleftmargin}{.125in}
\setlength{\bibindent}{-\bibleftmargin}
\bibliography{ref}

\normalsize
\section*{\LARGE Supplementary Information}
\subsection*{S-I\hspace*{8pt} Proof of Lemma 1:} Simple use of the total probability lemma yields:
\begin{equation}\tag{S1}
\bP(O|E)=\sum_{S\in Val(\bb S)}\bP(O|S,E)\bP(S|E).
\end{equation}
Equation (S1) immediately reveals a simple fact, namely, that $\bP(O|E)$ is a linear combination of the members of the set $\{\bP(O|S,E)\}_{S\in Val(\bb S)}$, an observation which grants the validity of the expression given in (\ref{eq_PLIF}) in the main text.

The key point which is left to be shown is the following: (Q.1) Why can the bounds given in (\ref{eq_PLIF}) be computed using the submodel retrieved in the process of obtaining the corresponding $\bb R_{\mc T}$ for the adopted IT $\mc T<p_l^\ast$? This is where the notion of PL comes into play. To articulate the intended line of reasoning let us introduce some notations first. According to Def. 3, any chosen IT $\mc T$ induces an IT-RS $\bb R_{\mc T}$. Let us partition the set of evidence variables $\bb E$ into three mutually disjoint sets $\bb E_{T}^+, \bb E_{T},$ and $\bb E_{T}^-$, where $\bb E_{T}$ denotes the set of variables in $\bb E$ which belong to the IT-RS $\bb R_{\mc T}$ (i.e., $\bb E_{T}:\triangleq\bb E\cap\bb R_{\mc T}$), $\bb E_{T}^+$ denotes the set of variables in $\bb E$ with PLs $\geq \mc T$, and finally, $\bb E_{T}^-$ denotes the set of  variables in $\bb E$ which are neither in $\bb E_{T}$ nor in $\bb E_{T}^+$ (i.e., $\bb E_{T}^-:\triangleq\bb E\setminus(\bb E_{T}\cup\bb E_{T}^+)$). Note that, by construction, the PLs of the variables in $\bb E_{T}^-$ are less than the adopted IT $\mc T$, hence the adopted notation. For example, for the setting depicted in Fig. \ref{fig_motive}(b) (corresponding to the IT $\mc T=p_l(\bb x)-\epsilon$), $\bb E_{T}=\varnothing, \bb E_{T}^+=\varnothing,$ and $\bb E_{T}^-=\{\bb y\}$. Also, for the setting depicted in Fig. \ref{fig_motive}(d) (corresponding to the IT $\mc T=p_l(\bb t_2)-\epsilon$), $\bb E_{T}=\{\bb y\}, \bb E_{T}^+=\varnothing,$ and $\bb E_{T}^-=\varnothing$. Next, we present a key result as a lemma.

\textbf{Lemma S.1.} Let $\bP(O|E)$ denote the posed causal query. For any chosen IT $\mc T<p_l^\ast$ and its corresponding IT-RS $\bb R_{\mc T}$, the following conditional independence relation holds: 
\begin{equation}\tag{S2}
(\bb O\ind\bb E_{T}^-|\bb R_{\mc T}\cup\bb E_{\mc T}^+).
\end{equation}

\textbf{Proof.} The relations between the PLs of the variables involved in the statement (S2) ensures that, according to $d$-separation criterion (Pearl, 1988), conditioning on the variables in $\bb R_{\mc T}\cup\bb E_{\mc T}^+$ blocks all the paths between the variables in $\bb O$ and $\bb E_{T}^-$, hence follows (S2).

The following two-part argument responds to the question posed in (Q.1) in the affirmative. First, notice that:
\begin{align}
\bP(O|S,E)\overset{}{=} \quad &\bP(O|S,E_{\mc T},E_{\mc T}^-,E_{\mc T}^+) \nonumber \\
\overset{}{=} \quad &\bP(O|R_{\mc T}, E_{\mc T}^-,E_{\mc T}^+) \nonumber \\
\overset{(S2)}{=} \quad &\bP(O|R_{\mc T},E_{\mc T}^+). \tag{S3}
\end{align}

Second, note that the process of obtaining $\bb R_{\mc T}$, namely, moving backwards from the variables in $\bb O\cup\bb E_{\mc T}^+$ until $\bb R_{\mc T}$ is reached, ensures that the submodel retrieved in this process suffices for the derivations of $\bP(O| R_{\mc T}, E_{\mc T}^+)$. Using the approach introduced in \cite{geiger1989d} for identifying the relevant information for the derivation of a query in a Bayesian network, this follows from the following fact: Conditioned on $\bb R_{\mc T}\cup\bb E_{\mc T}^+$, the set $\bb O$ is $d$-separated from all the nodes in the set $An(\bb O\cup\bb E)\setminus \bb R_{\mc T}$ whose PLs are less than the adopted IT $\mc T$. Note that $An(\bb O\cup\bb E)$ denotes the ancesteral graph for the nodes in $\bb O\cup\bb E$. This completes the proof. \hfill $\blacksquare$

\subsection*{S-II\hspace*{8pt} The Rationale behind Remark 1:}
Case (i) and Case (iii) immediately follow from Lemma 1 in the main text. Case (ii) implies that all the ancestors of variables in $\bb O\cup\bb E$ are retrieved, hence the sufficiency of the retrieved submodel for the exact derivation of the query; see also Sec. S-III.

\subsection*{S-III\hspace*{8pt} On the Special Case of Having $p_l^{\ast}=T_0$:}
In such circumstances, to derive $\bP(O|E)$, the set of all the ancestors of variables in $\bb O\cup\bb E$ should be retrieved and then inference should be carried out on the retrieved submodel.
\end{document}